\documentclass[runningheads]{llncs}
\usepackage{graphicx}
\usepackage{algorithm}
\usepackage{algorithmic}
\usepackage{amsmath}
\usepackage{amssymb}
\usepackage{graphicx}
\usepackage{array}
\usepackage{booktabs}
\usepackage{balance}
 \usepackage{mathtools}
\usepackage{wrapfig}

\usepackage{caption}
\usepackage{subcaption}
\usepackage{multirow}
\usepackage{multicol}
\usepackage{float}
\usepackage{enumerate}
\usepackage[shortlabels]{enumitem}
 \usepackage{color,soul}
\usepackage{tabularx}
\usepackage{setspace}
\usepackage{hyperref}
\newcommand{\repeatthanks}{\textsuperscript{\thefootnote}}
\newenvironment{bothindent}
{\par\leftskip1.5cm\relax\rightskip1.5cm\relax}
{\par\leftskip0cm\relax\rightskip0cm\relax}

\begin{document}
\title{``Let's Eat Grandma'': Does Punctuation Matter in Sentence Representation?}
\titlerunning{Let's Eat Grandma}
%
\author{Mansooreh Karami\thanks{Authors contributed equally to this work.}\orcidID{0000-0002-8168-8075} \and \\
Ahmadreza Mosallanezhad\repeatthanks\orcidID{0000-0003-1907-3536} \and \\
Michelle V Mancenido\orcidID{0000-0002-3000-8922} \and\\
Huan Liu\orcidID{0000-0002-3264-7904}}

%
\authorrunning{M. Karami et al.}
\institute{Arizona State University, Tempe AZ, USA
\email{\{mkarami,amosalla,mmanceni,huanliu\}@asu.edu}}
%
\maketitle              
\begin{abstract}
Neural network-based embeddings have been the mainstream approach for creating a vector representation of the text to capture lexical and semantic similarities and dissimilarities. In general, existing encoding methods dismiss the punctuation as insignificant information; consequently, they are routinely treated as a predefined token/word or eliminated in the pre-processing phase. However, punctuation could play a significant role in the semantics of the sentences, as in ``Let's eat\hl{,} grandma" and ``Let's eat grandma". We hypothesize that a punctuation-aware representation model would affect the performance of the downstream tasks.
Thereby, we propose a model-agnostic method that incorporates both syntactic and contextual information to improve the performance of the sentiment classification task. We corroborate our findings by conducting experiments on publicly available datasets and provide case studies that our model generates representations with respect to the punctuation in the sentence.

\end{abstract}
\section{Introduction}
\label{s:intro}

{\setstretch{0.75}
\begin{bothindent}

\textit{\footnotesize{According to a famous legend, Julius Caesar had decided to grant amnesty to one of his unscrupulous generals, who had been fated to be executed. ``Execute not, liberate,'' Caesar had ordered his guards. However, the message had been delivered with a small but calamitous error: ``Execute, not liberate.''\\}}
\end{bothindent}
}

\noindent
The recent paradigm shift to pre-training the NLP models with language modeling has gained tremendous success across a wide variety of downstream tasks. Word and sentence embeddings from these pre-trained language models have revolutionized the modern NLP research and reduced the non-trivial computational time of training NLP-related tasks. BERT~\cite{devlin2019bert}, an example of a pre-trained language model, addresses limitations of other methods by incorporating context from both directions to capture the semantic concepts more accurately~\cite{yenicelik2020does}.


In pre-trained language models, punctuation is often treated as an ordinary word or as a predefined token in the data or, in some cases, filtered out during the pre-processing phase~\cite{karami2021profiling,li2019generative,mosallanezhad2022domain}. The lack of considerable attention to punctuation in NLP models stems from the fact that punctuation has long been considered as cues that only aid text's readability, thus not providing additional semantic value to the sentence's coherence~\cite{ek2020does}.
However, studies show that the misplacement or elimination of these symbols can change the original meaning or obscure a text's implicit sentiment~\cite{altrabsheh2014sentiment,wang2014feature} as it conveys rich information about structural relations among the elements of a text. For example, ``No investments will be made over three years'' and ``No\hl{,} investments will be made over three years'' have drastically different meanings and implications. But BERT, as a representation tool, will assign a fixed predefined token to the punctuation treating it as an ordinary word in the data; under BERT, the vector representations of these two sentences are nearly the same. On the other hand, methods that account for punctuation are typically model-specific and cannot be integrated into SOTA representation models.


In this work, we hypothesize that trivializing the role of punctuation in \emph{sentiment analysis tasks} results in the degraded quality of representations which consequently, affects traditional measures of classifier performance. To provide evidence, we propose a model-agnostic module for representing the syntactic and contextual information that could be derived from punctuation. Our approach is based on an encoder that integrates structural and textual embedding to capture sentence-level semantics accurately through the use of parsing trees. Previous works on parsing trees have shown that there is an association between a text's punctuation and syntactic structure~\cite{li2019generative}.


\noindent
The following summarizes the major contributions of this work: 
\begin{itemize}
    \item We conduct preliminary experiments to show that the state-of-the-art representation learning models do not distinguish between sentences with and without punctuation~($\mathcal{x}$\ref{s:exp});
    \item We develop~($\mathcal{x}$\ref{s:proposed}) and evaluate~($\mathcal{x}$\ref{s:exp}) a model-agnostic methodology for sentiment analysis that augments the structure of the sentences to the original sentence embedding which can be integrated into SOTA representation models;
    \item We provide case studies to demonstrate that the proposed model yields proper representation for cases when punctuation change and do not change the meaning of the sentences~($\mathcal{x}$\ref{sec::case_study}).
\end{itemize}

\section{Related Work}
\label{s:related}
The proposed methodology spans the subject domains of word and sentence embeddings, punctuation in NLP tasks such as sentiment analysis, and tree-structured encoding. The current state-of-the-art in these areas is discussed in this section.  

\subsection{Embeddings}
Word and sentence embeddings are techniques used to map text data to vector representations so that the distance between the vectors corresponds to their semantic proximity. Word2vec had been applied on many tasks since it was introduced in 2013~\cite{mikolov2013efficient}. Although this neural network-based model could effectively encode the semantic and syntactic meaning of the text into vectors, word2vec is sub-optimal for syntax-based problems such as Part-of-Speech (POS) tagging or dependency parsing~\cite{ling2015two}. In recent years, embeddings such as BERT~\cite{devlin2019bert} improved on term-based embeddings by not only encoding the semantic information of words but also their contextualized meanings (i.e. terms and related contexts). Despite proving its usefulness across a wide range of tasks in NLP, BERT has been shown lacking in some aspects, such as common sense, pragmatic inferences, and the meaning of negations~\cite{ettinger2020bert}. 

One prevailing issue in sentiment analysis is that these representations typically fail to distinguish between words with similar contexts but opposite sentiment polarities (e.g., wonderful vs. terrible) because they were mapped to vectors that were closely contiguous in the latent space~\cite{zhang2018deep}. Thus, researchers proposed various word embedding methods to encode sentiments~\cite{lin2017structured,labutov2013re,maas2011learning,bespalov2011sentiment}. In this work, we propose a novel sentence embedding as an improvement over current methods for sentiment analysis tasks. 

\subsection{Punctuation in NLP}

Punctuation has long been considered the visual equivalent of spoken-language prosody, thus only providing cues that aid a text's readability. However, Nunberg~\cite{nunberg1990linguistics} argued that punctuation has a more important role. He defines it as a linguistic subsystem related to grammar that conveys rich information about the structural relations among the elements of a text~\cite{nunberg1990linguistics}. 

In NLP, the inclusion of punctuation marks has been shown to be useful in syntactic processing~\cite{lou2019neural} and could be used to enhance grammar induction in unsupervised dependency parsing. As an example, Spitkovsky et al.~\cite{spitkovsky2011punctuation} showed improved performance by splitting sentences at their punctuation to impose parsing restrictions over their fragments. Additionally, in the context of  sentiment analysis, punctuation marks have been shown to add extra value to the sentiment~\cite{altrabsheh2014sentiment,wang2014feature,pang2002thumbs} and could be used to create more meaningful syntax trees~\cite{li2019generative,agarwal2011sentiment}.



Despite evidence that incorporating punctuation improves aspects of an NLP's performance, very few NLP models make significant use of these symbols, which we concurrently address in the proposed methodology. Moreover, we investigate sentiments at the sentence-level.  

\subsection{Tree-Structured Encoders}
Tree-structured encoders, which have been shown to perform as well as their sequential counterparts, are representations constructed from the syntactic structure of groups of words or sentences. 
An example of a tree-structured encoder is the Tree-LSTM, a generalization of the long short term memory (LSTM) architecture that accounts for the topological structure of sentences ~\cite{tai2015improved}.  Each unit in the Tree-LSTM consists of values provided by the input vector and the hidden states of its children (as derived from the syntactic tree); in contrast, the standard LSTM only considers hidden states from the previous time step. Tree-LSTM was inspired by an RNN-based compositional model that captured the parent representation in syntactic trees~\cite{socher2011parsing,socher2013recursive}.

In addition to changing the LSTM architecture, another method to capture the syntactic structure of sentences is by directly using the LSTM architecture to code the syntactic structures. Liu et al.~\cite{liu2017structural} encoded the variable-length syntactic information, i.e. the path from leaf node to the root node in the constituency or dependency tree, into a fixed-length vector representation to embed the structural characteristics of the sentences on neural attention models for machine comprehension tasks. To jointly learn syntax and lexicon, Shen et al.~\cite{shen2018neural} proposed a Parsing-Reading-Predict neural language model (PRPN) that learns the syntactic structure from an unannotated corpus and uses the learned structure to form a premier language model. There has also been some work that extended the Transformer~\cite{vaswani2017attention} architecture for syntactic coding. 

The work in this paper augments constituency trees to the original word embedding to record the position of the punctuation by capturing structural information.

\section{Problem Statement}
\label{problem_statement}
Let $\mathcal{X} = \{ (\mathbf{x_1}, y_1), (\mathbf{x_2}, y_2), ..., (\mathbf{x_N}, y_N) \}$ denote a set of $N$ textual data with text $\mathbf{x_i}$ and the sentiment label $y_i$ for sample $i$. Each text $\mathbf{x_i}$ consists of 
sequence of words/punctuation $\mathbf{x_i} = [w_1 \; w_2 \; ... \; w_M]$, where 
$M$ represents the number of words and punctuation in the text.
Since the punctuation and their position affects the structure of the sentence and its meaning, we focus on generating a robust sentence embedding for sentiment analysis with respect to the structure of the sentence.
Formally, this problem can be stated as follows:
\begin{problem}
{\fontsize{9.95}{10}\selectfont
Given a set of textual data $\mathcal{X}$ comprising of words and punctuation, learn an embedding $\mathbf{E}$ which accounts for the constituency tree structure of the sentences and finds a function $\mathbf{F}$ for sentiment classification.}
\end{problem}

\section{Proposed Model}
\label{s:proposed}
\begin{figure}[h]
  \centering
  \includegraphics[width=\textwidth]{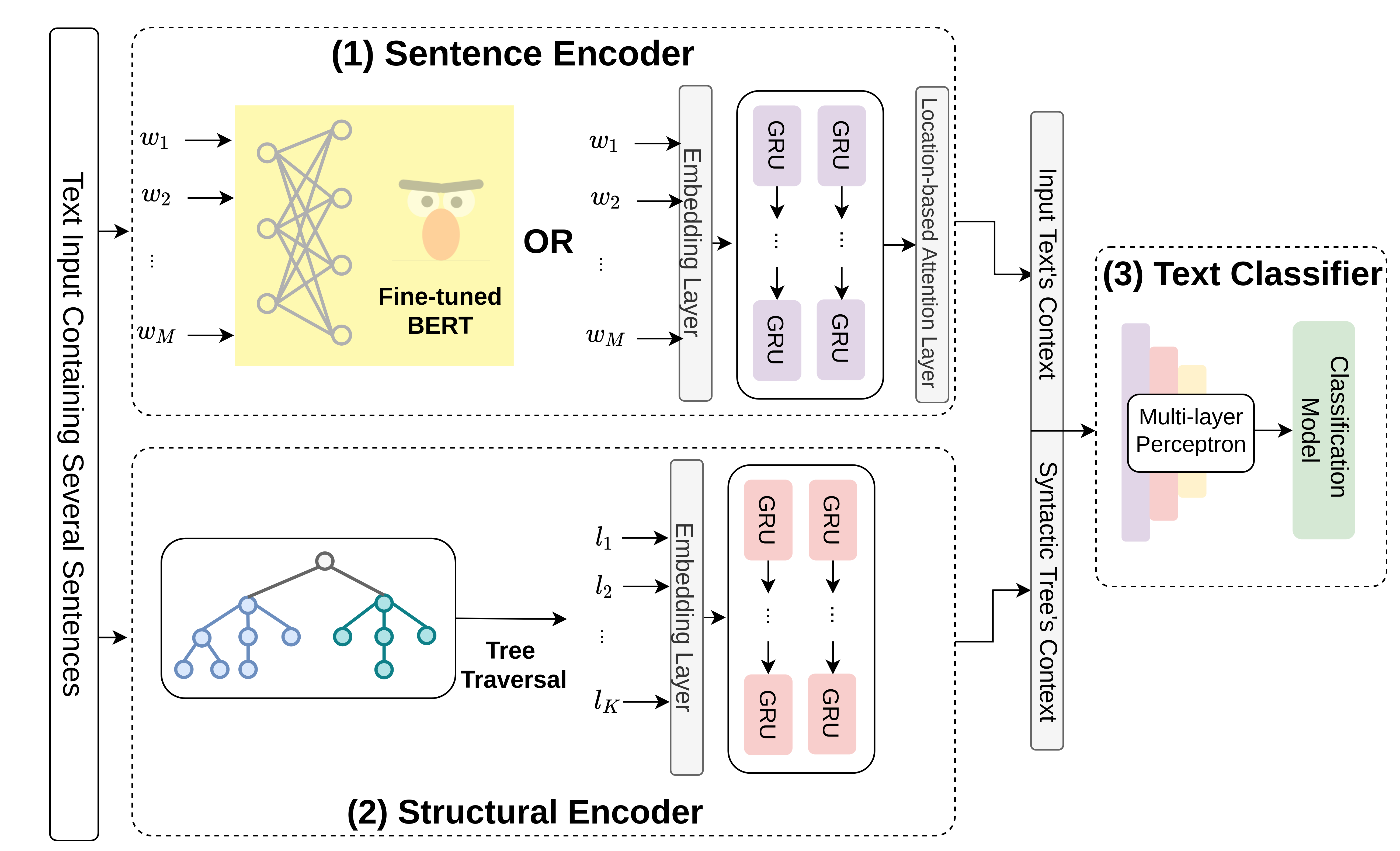}
  \caption{The three components of the model: (1) the sentence encoder that captures the input context, (2) the syntactic tree encoder which accounts for the structural content, and (3) the sentiment analysis classifier.}
  \label{fig::architecture} 
\end{figure}

We hypothesize that due to the effect of punctuation on the constituency structures of the sentences, adding the structural embedding of the sentences could improve the vector representation of sentences. The general framework of the proposed model is shown in Figure \ref{fig::architecture}. The proposed model has three major components: (1)~a sentence encoder, (2)~a structural encoder, and (3)~a text classifier. In the following discussion, we describe in detail the sentence and structural encoders and discuss how these two methods are integrated into a robust framework that improves embedding and classification performance. 

\subsection{Sentence Embedding for Sentiment Analysis}
In sentiment analysis, textual data is first converted into vectors or matrices. The ability of recurrent neural networks (RNNs) to model order-sensitive data makes it an effective choice for modeling textual data, where the order of words alter the contextual meaning. Our framework uses a bi-directional gated recurrent unit (BiGRU), an RNN that models contextual meanings more effectively than uni-directional networks~\cite{kiperwasser2016simple}. However, later as demonstrated in the experiments, we also considered a fine-tuned BERT instead of the BiGRU module in creating the text embeddings. This will ensure the generalization of our method for other representation models.

To create the text embeddings, a sample, $\mathbf{x}_i = [{\bf w}_1 \; {\bf w}_2 \; ... \; {\bf w}_M]$, is passed through an embedding layer which converts each word ${\bf w}_j$ to its representation. This layer has a tensor of dimension $|V| \times d_w$, where $V$ is the vocabulary and $d_w$ is the dimension of the word embeddings. The representations will be fed to a BiGRU that yields the following $M$ outputs:  
\begin{equation}
    (\overrightarrow{ \bf h}_m, \overleftarrow{\bf h}_m) = \text{BiGRU}({\bf w}_m, (\overrightarrow{\bf h}_{m-1}, \overleftarrow{\bf h}_{m-1}))
\end{equation}
where $\overrightarrow{\bf h}_m$ and $\overleftarrow{\bf h}_m$ are, respectively, the forward and backward outputs of the BiGRU at time step $m \in M$.
These BiGRU's outputs are then concatenated to form a fixed-length context vector:
\begin{equation}
    \mathbf{H}_m = \text{Concat}(\overrightarrow{\bf h}_m, \overleftarrow{\bf h}_m)
\end{equation}

Further, to establish a comprehensive context vector, an attention mechanism was included by augmenting a location-based attention layer~\cite{luong2015effective}. The weighted average of the importance values $a_m \in \mathbf{H_m}$ provided by the attention layer creates the final context vector:
\begin{equation}
    \mathbf{H'} = \sum_i a_i \mathbf{H}_i
\end{equation}

Using the context vector $\mathbf{H'}$ with a Multi-Layer Perceptron (MLP) classifier yields good performance on sentiment analysis tasks~\cite{sachin2020sentiment,yang2016hierarchical}.

Information learned from BiGRU/BERT, as described in this subsection, will be combined with the encoded syntactic structure of the sentence. This will enhance the context vector to include salient information provided by punctuation.  

\subsection{Enhanced Embedding}

\begin{figure*}[h]
    \centering
    \includegraphics[width=0.75\textwidth]{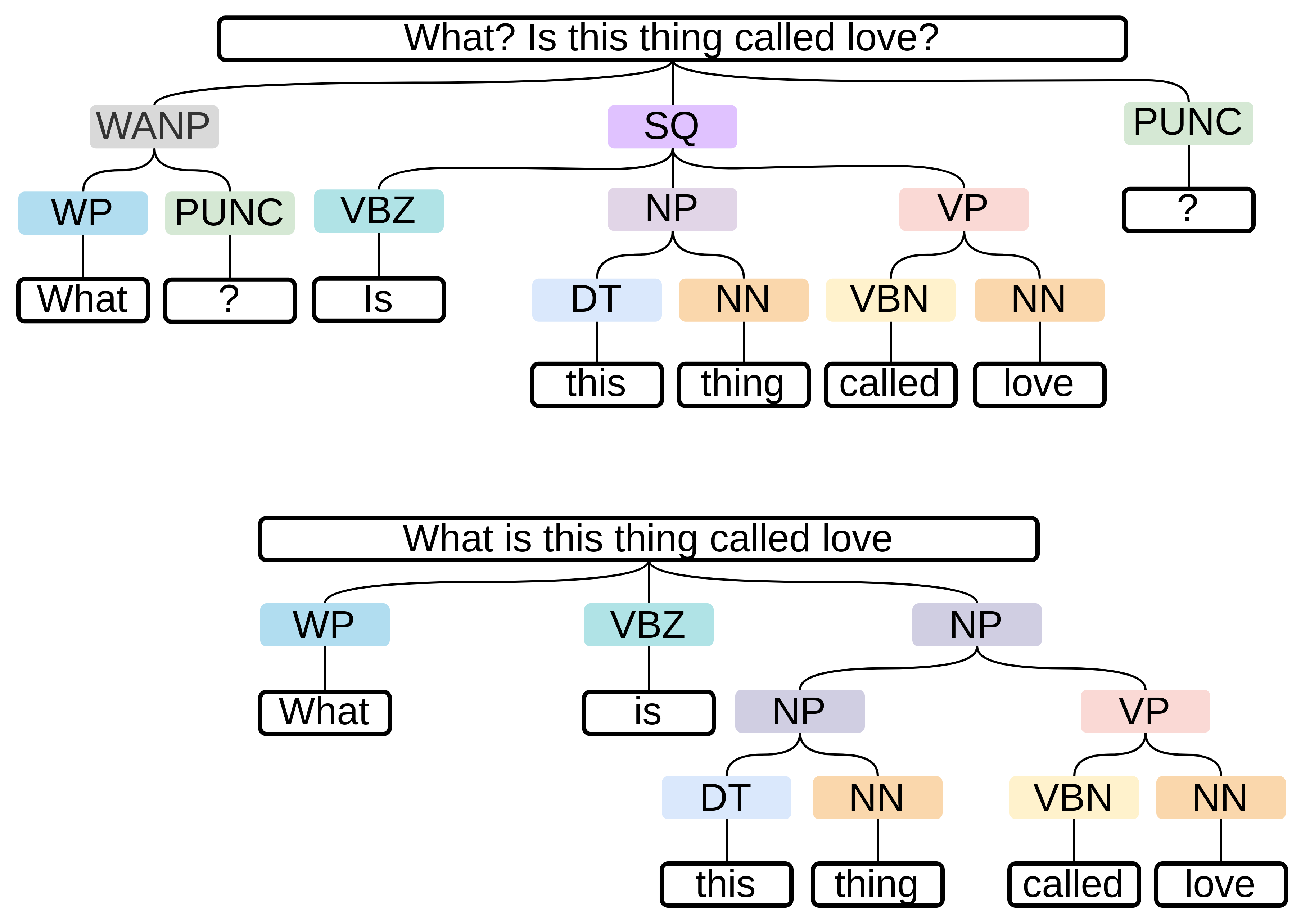}
     
    \caption{The constituency tree of a text with and without punctuation, ``what is this thing called love'' versus ``what? is this thing called love?''}    \label{fig::constituency_tree} 
\end{figure*}

We use a constituency tree to analyze sentence structure and organize words into nested constituents. In the constituency tree, words are represented by the leaves while the internal nodes show the phrasal (e.g. S, NP and VP) or pre-terminal Part-Of-Speech (POS) categories. Edges in the tree indicate the set of grammar rules. Figure~\ref{fig::constituency_tree} shows an example of a constituency tree that demonstrates the parsing of a sample sentence. 
Subsequent to the generation of the syntactic tree, we adopt the word-level approach in Liu et al. ~\cite{liu2017structural} to capture syntactic information but in a sentence-level manner. We use the traversal of the syntactic tree $T$ to pass it through a bi-directional GRU and create a representation of $T$. Because the order of the nodes in a tree impact the traversal result, we use BiGRU to create a correct representation:
\begin{equation}
    (\overrightarrow{\bf h}_t, \overleftarrow{\bf h}_t) = \text{BiGRU}({\bf l}_t, (\overrightarrow{\bf h}_{t-1}, \overleftarrow{\bf h}_{t-1}))
\end{equation}
where ${\bf l}$ is the value of the tree node and ${\bf h}_t$ shows the BiGRU output. We consider the last output of the BiGRU, $\mathbf{H}_T = \text{Concat}(\overrightarrow{\bf h}_t, \overleftarrow{\bf h}_t)$, as the context of the syntactic tree. 

Finally, to balance the effect of the extracted contexts, the context of the text $\mathbf{H'}$ and the context of its syntactic tree $\mathbf{H}_T$ are passed through a feed-forward neural network to create the enhanced text representation:

\begin{equation}
    \mathbf{H_F} = \text{MLP}(\mathbf{H'}, \mathbf{H}_T)
\end{equation}

where $\mathbf{H_F}$ is the enhanced text representation containing the text's semantic information and information about its syntactic tree. This enhanced representation could now be used for sentiment analysis tasks.

\section{Experimental Settings}
In this section, we present details about the datasets, the implementation details, as well as the baseline methods used for the experiments. 
\label{s:exp}

\subsection{Datasets}

Four publicly available datasets -- IMDB, Rotten Tomatoes (RT), Stanford Sentiment Treebank (SST), and Yelp Polarity (Yelp P.) -- were used to evaluate and compare the proposed method with other baselines. The IMDB movie reviews dataset contains $50,000$ movie reviews, with each review labeled as `positive' or `negative'. In a similar fashion, the Rotten Tomatoes dataset contains $480,000$ movie reviews from the Rotten Tomatoes website, labeled as `fresh' (positive) or `rotten' (negative). As a more challenging task, we consider the SST-2 dataset, which consists of $10,754$ samples having a binary label of positive and negative sentiment. Finally, we utilized a subsample of $100,000$ reviews from Yelp Polarity dataset which uses `negative' and `positive' labels instead of the five point star scale~\cite{zhang2015character}. Table~\ref{datatable} summarizes some key statistics of each dataset. We used 10-fold cross-validation with 45/5/50 for Train/Validation/Test split configuration to compare the proposed model with other baselines.

\begin{table}
\centering
\caption{The statistics of the datasets.}
\label{datatable}
\small
\begin{tabular}{lp{2mm}cp{2mm}cp{2mm}c} 
\toprule
\multicolumn{1}{c}{\textbf{Dataset}}&   & \textbf{\# of Samples} && \begin{tabular}[c]{@{}c@{}}\textbf{Avg Text Length }\\\textbf{(\# of Words)}\end{tabular} && \textbf{\# of Sentences}  \\ 
\midrule
IMDB                     &            & 50,000         &        & 231.1 $\pm$ 171.3                                                       &                  & 536,641                   \\
RT                                 &  & 480,000              &  & 21.8 $\pm$ 9.3                                                                       &     & 601,787                   \\
SST-2                             &   & 10,754               &  & 19.4 $\pm$ 9.3                                                                          &  & 11,855                    \\
Yelp P.                            &  & 100,000              &  & 133 $\pm$ 122.5                                                                         &  & 814,596                   \\
\bottomrule
\end{tabular}
\end{table}



\subsection{Implementation Details}
In this subsection, we discuss the parameters and implementation details of the proposed model for conducting the experiments\footnote{The code for this work is available at: \href{https://github.com/mansourehk/Grandma}{https://github.com/mansourehk/Grandma}}. Based on the average number of the words in the datasets (Table~\ref{datatable}), we truncate every textual data to $128$ words.
Next, we extract the syntactic tree for each sentence, in the spirit of Liu et al.~\cite{liu2017structural} but in a sentence-level manner using Spacy toolkit\footnote{Available at https://spacy.io/}. Finally, to combine all trees related to a text, an empty root was added as the parent of all the other roots of the syntactic trees. Children are arranged based on the order of the sentences in the text (Figure~\ref{fig::architecture}). 

We use GloVe 100d~\cite{pennington2014glove} to replace each word with its corresponding word vector to convert sentences into matrices. For words and POS tags that are not included in GloVe, a trainable random vector was used as a proxy. We use a 1-layer BiGRU with $256$ hidden neurons to generate the text's context vector and a $128$-hidden neuron BiGRU for the syntactic tree's context vector. To combine both context vectors, we use a simple neural network with $512$ output neurons. The output of this neural network is the final context vector $\mathbf{H_F}$ containing both semantic and syntactic information of the input text:
\begin{align}
    \mathbf{o} &= \textrm{tanh} ( \mathbf{W}_{F}^{(1)} \mathbf{ (H'||H_T) } + \mathbf{b}_{F}^{(1)} ) \\
    \mathbf{H_F} &= \textrm{tanh}(\mathbf{W}_{F}^{(2)} \mathbf{o} + \mathbf{b}_{F}^{(2)}),
\end{align}
where $||$ is the concatenation operator, $(\mathbf{W, b})$ are the learnable weights, and $\mathbf{H'}$, $\mathbf{H_T}$ are the input's context and the syntactic tree's context vectors, respectively.

The integrated context vector $\mathbf{H_F}$ is used for text classification. The neural network classifier includes three layers with $512$, $128$, and $C$ number of neurons, respectively, where $C$ is the number of classes. Model parameters $\theta$ and the data labels $y$ are updated using a cross-entropy loss function in the training phase:
\begin{equation}
    L(\theta) = -\frac{1}{N} \sum_{i=1}^{N} \sum_{j=1}^{C} y_{ij} \log(p_{ij}),
\end{equation}
where $N$ is the number of samples. We use the Adam optimizer~\cite{kingma2014adam} to update the parameters of the network. 

\subsection{Baseline Methods}

Several embedding methods are implemented to generate sentence representations for comparison with the proposed model. The vectors created by these sentence encoders are used as inputs to the three-layered neural network classifier. Each sentence representation method is described below.
\begin{itemize}
    \item \textbf{BERT}~\cite{devlin2019bert}: Bidirectional Encoder Representations from Transformers is a model used for various NLP tasks, including sentiment analysis. In this paper, a pre-trained base BERT is used to extract the sentence embeddings.
    
    \item \textbf{BiGRU}: similar to the approach in Mosallanezhad et al.~\cite{mosallanezhad2019deep}, we design a baseline that uses a bidirectional GRU to create a context vector based on the input text. Each word is replaced by its corresponding GloVe vector and passed through a bidirectional Gated Recurrent Unit. The final output of the BiGRU is then considered as the context vector.

    \item \textbf{BiGRU+Attn}: similar to the BiGRU method, but uses a location-based attention layer~\cite{luong2015effective} to create the context vector.

    \item \textbf{SEDT-LSTM}~\cite{liu2017structural}: creates a word-level embedding by including the dependency tree of the sentences. For each word $w$ in the text, this method merges the GloVe vector of $w$ with the fixed-length context vector extracted from the dependency tree. To create this context vector, all the words in the path from $w$ to the root node in the dependency tree are fed to an LSTM.
\end{itemize}

We integrated our model-agnostic module (i.e., the syntactic tree encoder) to the BiGRU, BiGRU+Attn, and BERT.

\section{Discussion and Experimental Results}

In this section, we conduct experiments to evaluate the effectiveness of our method in sentiment analysis tasks. We propose two major research questions:

\begin{enumerate}[leftmargin=.5in, label={(\bfseries Q\arabic*)}]
        \item How do other methods behave in terms of the embeddings and perform in terms of the sentiment classification task when punctuation is included in the input text?
        \item How well does the proposed method incorporate the effect of the punctuation in the sentence embeddings?
\end{enumerate}


\begin{figure}[h]
    \centering
         \includegraphics[width=0.8\textwidth]{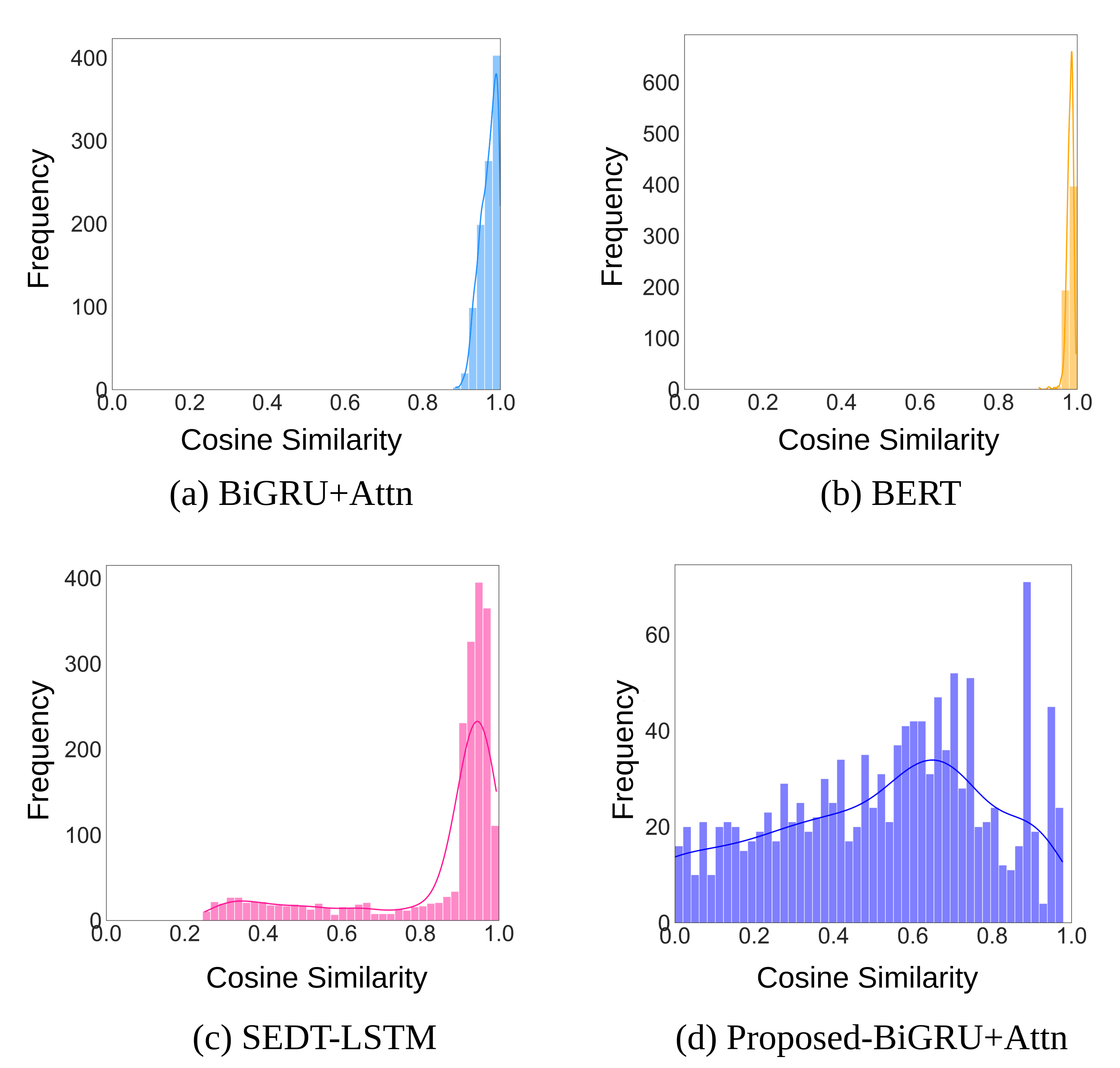}
    \caption{The histogram of cosine similarities between sentence embeddings with and without punctaution. Higher similarity means that the embeddings are close to each other.}
    \label{fig:embs_before}
\end{figure}

Figure~\ref{fig:embs_before} shows the similarity between sentence embeddings with and without punctuation in the text. To calculate the similarity between embeddings, we use the cosine similarity measure:

\begin{equation}
    \text{CosineSim}(\mathbf{E}_w, \mathbf{E}_{wo}) = \frac{\mathbf{E}_w \cdot \mathbf{E}_{wo}}{||\mathbf{E}_w|| || \mathbf{E}_{wo}||},
\end{equation}
where $\mathbf{E}_w$ and $\mathbf{E}_{wo}$ are the sentence embeddings with and without punctuation, respectively. The cosine similarity measure is close to $1.0$ when context vectors are close to each other.

\textbf{Q1.}
In Figure~\ref{fig:embs_before} (a-b), it is observable that BERT and Recurrent Neural Networks (BiGRU+Attn) have higher cosine similarity measures, implying that they do not produce different embeddings for sentences with and without punctuation. The minimum similarity between embeddings for these models is approximately $0.9$. This finding corroborates our hypothesis that these models \textit{consider punctuation as just another fixed word/token in the data}, strongly justifying the development of an enhanced representation method.

Additionally, 
Table~\ref{table:res_wopunc} shows the accuracy of the baseline models on the aforementioned datasets when punctuation is excluded. By comparing it with the first section of Table~\ref{table:res_wpunc}, it is evident that the performance of the baselines are \textit{agnostic to the use of punctuation} due to their similar representation vectors in both cases. For the baselines, inclusion of the punctuation is almost irrelevant and even lowers performance in some cases providing evidence why most researchers exclude punctuation in the preprocessing phase. 

\begin{table}[h]
\centering
\small
\caption{\textit{Without Punctiation:} Performance (accuracy) of the baseline models on the datasets.}
\begin{tabular}{p{2.5cm}cp{2mm}cp{2mm}cp{2mm}c}
    \toprule
        & \multicolumn{7}{c}{\bf Datasets}   \\ \midrule
{\bf Model}  & {\bf IMDB} && {\bf RT} && {\bf SST-2}&& {\bf Yelp P.}  \\
\midrule
BiGRU        & 88.0 &&     69.1   &&    86.9 &&  84.8 \\
BiGRU+Attn   & 88.8 &&     70.0   &&    87.4  && 84.6 \\
BERT         & 92.3 &&     71.6   &&  91.7  && 90.6 \\ \bottomrule
\end{tabular}

\label{table:res_wopunc}
\end{table}


\textbf{Q2.} Figure~\ref{fig:embs_before} (c-d) shows the trend of cosine similarity when the syntactic information is augmented with the word embedding. The lower similarity values, ranging from as low as $0.10$ to only as high as $0.90$, indicate that the representation vectors of sentences with and without punctuation are distinct. While the SEDT-LSTM model shows promising results, the proposed model still outperforms SEDT-LSTM in the sentiment analysis task~(Table~\ref{table:res_wpunc}).
This difference is due to the fact that our model operates in \textit{a sentence-level manner and provides a richer structural embedding}, while SEDT-LSTM works as a word-level approach and does not account for the whole structure of the syntactic tree.

\begin{table}[h]
\centering
\small
\caption{\textit{With Punctuation:} Performance (accuracy) of SEDT-LSTM and our added module to different representation baselines when punctuation is included.}
\begin{tabular}{p{3.5cm}cp{2mm}cp{2mm}cp{2mm}c}
    \toprule
        & \multicolumn{7}{c}{\bf Datasets}   \\ \midrule
{\bf Model} & {\bf IMDB} && {\bf RT} && {\bf SST-2} && {\bf Yelp P.} \\
\midrule

{\fontsize{8.25}{8.5}\selectfont BiGRU}        & $88.1$  &&   $70.3$   &&  $87.3$ && $84.8$ \\
{\fontsize{8.25}{8.5}\selectfont BiGRU+Attn}   & $88.2$  &&   $70.5$   &&  $88.1$  && $84.8$\\
{\fontsize{8.25}{8.5}\selectfont BERT}         &  $92.1$  &&   $71.5$  &&   $91.7$   &&  $90.6$ \\ \midrule
{\fontsize{8.25}{8.5}\selectfont SEDT-LSTM}    & 91.1  &&   72.0   &&    90.5  && 85.1 \\
{\fontsize{8.25}{8.5}\selectfont Proposed-BiGRU} & 92.7 && 74.2 && 90.1 && 87.1 \\
{\fontsize{8.25}{8.5}\selectfont Proposed-BiGRU+Attn}     & 93.0&&    74.3   && 91.3 && 88.3\\
{\fontsize{8.25}{8.5}\selectfont Proposed-BERT}     & \textbf{94.6}&&    \textbf{74.8}   && \textbf{92.4} && \textbf{91.7} \\\bottomrule 
\end{tabular}

\label{table:res_wpunc}
\end{table}

\begin{table}[H]
\centering
\footnotesize
\caption{Examples of sentences in which punctuation changes the meaning of the sentence. The proposed method distinguishes between the two versions, with and without punctuation. In this experiment, we use both inputs on a single model.}
\label{tab:examples_sim1}
\begin{tabular}{cp{3.8cm}p{1mm}p{3.6cm}cccc} 
\toprule
\multicolumn{4}{c}{\begin{tabular}[c]{@{}c@{}}\textbf{ Examples in which Punctuation Change}\\\textbf{the Meaning of the Sentence }\end{tabular}} & \multicolumn{4}{c}{\textbf{ Cosine Similarity }}  \\ 
\midrule
  & \scriptsize{With Punctuation}                                    && \scriptsize{Without Punctuation}                                                                     & \scriptsize{Proposed} & \scriptsize{SEDT-LSTM} & \scriptsize{BERT} & \scriptsize{BiGRU+Attn}          \\ 
\midrule
1 & Now, my friends, listen to me.                      && Now my friends listen to me                                                             & 0.56     & 0.67      & 0.97 & 0.95                \\
2 & Help. wanted.                                       && Help wanted                                                                             & 0.51     & 0.45      & 0.99 & 0.99                \\
3 & What? Is this thing called \`{}love'?               && What is this thing called love                                                          & 0.75     & 0.78      & 0.98 & 0.99                \\
4 & No, investments will be made in United States       && No investments will be made in United States                                            & 0.57     & 0.55      & 0.96 & 0.96                \\
5 & If you go, pack your knitting needles.              && If you go pack your knitting needles                                                    & 0.43     & 0.67      & 0.97 & 0.98                \\
6 & When the plot kicks in, the film loses credibility. && When the plot kicks in the film loses credibility                                       & 0.48     & 0.78      & 0.96 & 0.94                \\
\bottomrule
\end{tabular}
\end{table}

\section{Case Studies}
\label{sec::case_study}
The cosine similarity of several sentences were also calculated to investigate how the methods compare when punctuation is removed. We combined the IMDB and Rotten Tomatoes datasets into a larger dataset, which is justifiable due to the similarity in the purpose and structure of the two datasets. The combined dataset was used to train the proposed model and the baseline methods. 

\begin{table}[H]
\centering
\footnotesize
\caption{The cosine similarity of sentences with and without punctuation in which the punctuation do not change the meaning of the sentence using different embedding methods. The proposed method can incorporate the syntactic tree's information better than the baselines. In this experiment, we use both inputs on a single model.}
\label{tab:examples_sim2}
\begin{tabular}{cp{3.5cm}p{1mm}p{3.5cm}cccc} 
\toprule
\multicolumn{4}{c}{\begin{tabular}[c]{@{}c@{}}\textbf{Examples in which Punctuation do not Change}\\\textbf{the Meaning of the Sentence }\end{tabular}} & \multicolumn{4}{c}{\textbf{ Cosine Similarity }}  \\ 
\midrule
  & \scriptsize{With Punctuation}                                    && \scriptsize{Without Punctuation}                                                                     & \tiny{Proposed} & \tiny{SEDT-LSTM} & \tiny{BERT} & \tiny{BiGRU+Attn}          \\ 
\midrule
        7 & A gorgeously strange movie, heaven is deeply concerned with morality, but it refuses to spell things out for viewers. && A gorgeously strange movie heaven is deeply concerned with morality but it refuses to spell things out for viewers & 0.89 & 0.91 & 0.98 & 0.99 \vspace{4pt}\\
        
        8 & But, like silence, it's a movie that gets under your skin. && But like silence its a movie that gets under your skin & 0.96 & 0.98 & 0.98 & 0.99 \vspace{4pt}\\
        
        9 & You will be required to work twenty-four hour shifts. && You will be required to work twenty four hour shifts & 0.99 & 0.99 & 0.99 & 0.99 \vspace{4pt}\\
        
        10 & The talents of the actors helps ``Moonlight Mile'' rise above its heart-on-its-sleeve writing. && The talents of the actors helps Moonlight Mile rise above its heart on its sleeve writing & 0.97 & 0.95 & 0.97 & 0.98 \vspace{4pt}\\

        11 & It's a fine, old - fashioned - movie. movie, which is to say it's unburdened by pretensions to great artistic significance. && It s a fine old fashioned movie movie which is to say it s unburdened by pretensions to great artistic significance & 0.95 & 0.94 & 0.98 & 0.99 \vspace{4pt}\\
        
        12 & Her favorite pies were lemon meringue, apple, and pecan. && Her favorite pies were lemon meringue apple and pecan & 0.83 & 0.93 & 0.98 & 0.97 \vspace{4pt}\\
\bottomrule
\end{tabular}
\end{table}

Table~\ref{tab:examples_sim1}-\ref{tab:examples_sim3} shows the cosine similarity measures of sample sentences with and without punctuation for all models. What is interesting in the results is that \textit{the proposed model clearly distinguishes the syntax between sentences where punctuation is necessary} (similarity measures are lower). Specifically, this is apparent in sentences provided in Table~\ref{tab:examples_sim1}.

Further, \textit{if the context of the sentence is agnostic with respect to the punctuation, our proposed model still performs relatively well} (high cosine similarity measure). This is evident in sentences 7-11 in Table~\ref{tab:examples_sim2}. In a specific example, sentence 12 shows a case where the Oxford/serial comma helps in preventing ambiguity. Without the serial comma, `apple and pecan' could be interpreted as a pie containing both apples and pecans. By looking into the cosine similarities, the proposed method seems to distinguish this nuance.

Additionally, to confirm our hypothesis that baselines such as BERT do not differentiate among different kinds of punctuation, we randomly replaced the punctuation in sentences with other types. It is evident from Table~\ref{tab:examples_sim3} results that \textit{the proposed method creates different representations when punctuation changes} while BERT and BiGRU provided nearly similar representations.

\begin{table}[h]
\centering
\footnotesize
\caption{Examples of sentences with random punctuation alongside their cosine similarity using different embedding methods. The proposed method can incorporate the syntactic tree's information better than the baselines. In this experiment, we use both inputs on a single model.}
\label{tab:examples_sim3}
\begin{tabular}{cp{3.5cm}p{1mm}p{3.5cm}cccc} 
\toprule
\multicolumn{4}{c}{\begin{tabular}[c]{@{}c@{}}\textbf{Examples in which Random Punctuation May}\\\textbf{Change the Meaning of the Sentence }\end{tabular}} & \multicolumn{4}{c}{\textbf{ Cosine Similarity }}  \\ 
\midrule
  & \scriptsize{With Punctuation}                                    && \scriptsize{Without Punctuation}                                                                     & \tiny{Proposed} & \tiny{SEDT-LSTM} & \tiny{BERT} & \tiny{BiGRU+Attn}          \\ 
\midrule
        13 & Now\hl{,} my friends\hl{,} listen to me\hl{.} && Now\hl{.} my friends\hl{!} listen to me\hl{,} & 0.52 & 0.59 & 0.96 & 0.98 \vspace{4pt}\\
        14 & Help\hl{.} wanted\hl{.} && Help\hl{,} wanted\hl{?} & 0.67 & 0.61 & 0.96 & 0.98 \vspace{4pt}\\
        15 & What\hl{?} Is this thing called \hl{'}love\hl{'?} && What\hl{.} Is this thing called \hl{'}love\hl{'!} & 0.82 & 0.78 & 0.99 & 0.99 \vspace{4pt}\\
        16 & A gorgeously strange movie\hl{,} heaven is deeply concerned with morality\hl{,} but it refuses to spell things out for viewers\hl{.} && A gorgeously strange movie\hl{?} heaven is deeply concerned with morality\hl{.} but it refuses to spell things out for viewers\hl{,} & 0.91 & 0.94 & 0.98 & 0.99 \vspace{4pt}\\
        17 & But\hl{,} like silence\hl{,} it\hl{'}s a movie that gets under your skin\hl{.} && But\hl{!} like silence\hl{.} it\hl{?}s a movie that gets under your skin\hl{?} & 0.74 & 0.77 & 0.97 & 0.98 \vspace{4pt}\\
        18 & You will be required to work twenty\hl{-}four hour shifts\hl{.} && You will be required to work twenty\hl{!}four hour shifts\hl{,} & 0.94 & 0.95 & 0.99 & 0.99 \vspace{4pt}\\
\bottomrule
\end{tabular}
\end{table}

\section{Conclusion and Future Work}
\label{s:conc}

In this paper, we proposed a model-agnostic methodology for sentence embeddings that consider punctuation as a salient feature of textual data. By leveraging on the association between punctuation and syntactic trees, our model yielded embeddings that were consistently able to convey the contextual meaning of sentences more accurately. We integrate our proposed module into state-of-the-art representation models, including BERT, the gold standard for NLP tasks. The proposed model in this paper outperformed the baselines in distinguishing between sentences with and without punctuation, especially those that require punctuation to be sensical. Moreover, as task performance, it performed accurately on classifying opinions for the IMDB, Rotten Tomatoes, SST-2, and Yelp P. datasets. A possible direction for future research is to use syntactic trees for other NLP-related tasks, such as automated chatbots and machine comprehension. 
\section*{ACKNOWLEDGMENT}
The authors would like to thank Sarath Sreedharan (ASU) and Sachin Grover (ASU) for their comments on the manuscript.  This material is, in part, based upon works supported by ONR (N00014-21-1-4002) and the U.S. Department of Homeland Security (17STQAC00001-05-00)\footnote{Disclaimer: "The views and conclusions contained in this document are those of the authors and should not be interpreted as necessarily representing the official policies, either expressed or implied, of the U.S. Department of Homeland Security."}.
%
%
%
%
\bibliographystyle{splncs04}
\bibliography{references}

\begin{thebibliography}{10}
\providecommand{\url}[1]{\texttt{#1}}
\providecommand{\urlprefix}{URL }
\providecommand{\doi}[1]{https://doi.org/#1}

\bibitem{agarwal2011sentiment}
Agarwal, A., Xie, B., Vovsha, I., Rambow, O., Passonneau, R.J.: Sentiment
  analysis of {T}witter data. In: Proceedings of the workshop on language in
  social media (LSM 2011). pp. 30--38 (2011)

\bibitem{altrabsheh2014sentiment}
Altrabsheh, N., Cocea, M., Fallahkhair, S.: Sentiment analysis: towards a tool
  for analysing real-time students feedback. In: 2014 IEEE 26th international
  conference on tools with artificial intelligence. pp. 419--423. IEEE (2014)

\bibitem{bespalov2011sentiment}
Bespalov, D., Bai, B., Qi, Y., Shokoufandeh, A.: Sentiment classification based
  on supervised latent n-gram analysis. In: Proceedings of the 20th ACM
  international conference on Information and knowledge management. pp.
  375--382 (2011)

\bibitem{devlin2019bert}
Devlin, J., Chang, M.W., Lee, K., Toutanova, K.: {BERT}: Pre-training of deep
  bidirectional transformers for language understanding. In: NAACL-HLT (1)
  (2019)

\bibitem{ek2020does}
Ek, A., Bernardy, J.P., Chatzikyriakidis, S.: How does punctuation affect
  neural models in natural language inference. In: Proceedings of the
  Probability and Meaning Conference (PaM 2020). pp. 109--116 (2020)

\bibitem{ettinger2020bert}
Ettinger, A.: What {BERT} is not: Lessons from a new suite of psycholinguistic
  diagnostics for language models. Transactions of the Association for
  Computational Linguistics  \textbf{8},  34--48 (2020)

\bibitem{karami2021profiling}
Karami, M., Nazer, T.H., Liu, H.: Profiling fake news spreaders on social media
  through psychological and motivational factors. In: Proceedings of the 32nd
  ACM Conference on Hypertext and Social Media. pp. 225--230 (2021)

\bibitem{kingma2014adam}
Kingma, D.P., Ba, J.: Adam: A method for stochastic optimization. In: ICLR
  (Poster) (2015)

\bibitem{kiperwasser2016simple}
Kiperwasser, E., Goldberg, Y.: Simple and accurate dependency parsing using
  bidirectional {LSTM} feature representations. Transactions of the Association
  for Computational Linguistics  \textbf{4},  313--327 (2016)

\bibitem{labutov2013re}
Labutov, I., Lipson, H.: Re-embedding words. In: Proceedings of the 51st Annual
  Meeting of the Association for Computational Linguistics (Volume 2: Short
  Papers). pp. 489--493 (2013)

\bibitem{li2019generative}
Li, X.L., Wang, D., Eisner, J.: A generative model for punctuation in
  dependency trees. Transactions of the Association for Computational
  Linguistics  \textbf{7},  357--373 (2019)

\bibitem{lin2017structured}
Lin, Z., Feng, M., Santos, C.N.d., Yu, M., Xiang, B., Zhou, B., Bengio, Y.: A
  structured self-attentive sentence embedding. In: International Conference on
  Learning Representations (ICLR) (2017)

\bibitem{ling2015two}
Ling, W., Dyer, C., Black, A.W., Trancoso, I.: Two/too simple adaptations of
  word2vec for syntax problems. In: Proceedings of the 2015 Conference of the
  North American Chapter of the Association for Computational Linguistics:
  Human Language Technologies. pp. 1299--1304 (2015)

\bibitem{liu2017structural}
Liu, R., Hu, J., Wei, W., Yang, Z., Nyberg, E.: Structural embedding of
  syntactic trees for machine comprehension. In: Proceedings of the 2017
  Conference on Empirical Methods in Natural Language Processing. pp. 815--824
  (2017)

\bibitem{lou2019neural}
Lou, P.J., Wang, Y., Johnson, M.: Neural constituency parsing of speech
  transcripts. In: NAACL-HLT (1) (2019)

\bibitem{luong2015effective}
Luong, T., Pham, H., Manning, C.D.: Effective approaches to attention-based
  neural machine translation. In: EMNLP (2015)

\bibitem{maas2011learning}
Maas, A., Daly, R.E., Pham, P.T., Huang, D., Ng, A.Y., Potts, C.: Learning word
  vectors for sentiment analysis. In: Proceedings of the 49th annual meeting of
  the association for computational linguistics: Human language technologies.
  pp. 142--150 (2011)

\bibitem{mikolov2013efficient}
Mikolov, T., Chen, K., Corrado, G., Dean, J.: Efficient estimation of word
  representations in vector space. In: Bengio, Y., LeCun, Y. (eds.) 1st
  International Conference on Learning Representations, {ICLR} 2013,
  Scottsdale, Arizona, USA, May 2-4, 2013, Workshop Track Proceedings (2013)

\bibitem{mosallanezhad2019deep}
Mosallanezhad, A., Beigi, G., Liu, H.: Deep reinforcement learning-based text
  anonymization against private-attribute inference. In: Proceedings of the
  2019 Conference on Empirical Methods in Natural Language Processing and the
  9th International Joint Conference on Natural Language Processing
  (EMNLP-IJCNLP). pp. 2360--2369 (2019)

\bibitem{mosallanezhad2022domain}
Mosallanezhad, A., Karami, M., Shu, K., Mancenido, M.V., Liu, H.: Domain
  adaptive fake news detection via reinforcement learning. In: Proceedings of
  the ACM Web Conference 2022. pp. 3632--3640 (2022)

\bibitem{nunberg1990linguistics}
Nunberg, G.: The Linguistics of Punctuation. Center for the Study of Language
  (CSLI) (1990)

\bibitem{pang2002thumbs}
Pang, B., Lee, L., Vaithyanathan, S.: Thumbs up? sentiment classification using
  machine learning techniques. In: In proceedings of EMNLP (2002)

\bibitem{pennington2014glove}
Pennington, J., Socher, R., Manning, C.D.: {GloVe}: Global vectors for word
  representation. In: Proceedings of the 2014 conference on empirical methods
  in natural language processing (EMNLP). pp. 1532--1543 (2014)

\bibitem{sachin2020sentiment}
Sachin, S., Tripathi, A., Mahajan, N., Aggarwal, S., Nagrath, P.: Sentiment
  analysis using gated recurrent neural networks. SN Computer Science
  \textbf{1}(2),  1--13 (2020)

\bibitem{shen2018neural}
Shen, Y., Lin, Z., Huang, C.w., Courville, A.: Neural language modeling by
  jointly learning syntax and lexicon. In: International Conference on Learning
  Representations (2018)

\bibitem{socher2011parsing}
Socher, R., Lin, C.C., Manning, C., Ng, A.Y.: Parsing natural scenes and
  natural language with recursive neural networks. In: Proceedings of the 28th
  international conference on machine learning (ICML-11). pp. 129--136 (2011)

\bibitem{socher2013recursive}
Socher, R., Perelygin, A., Wu, J., Chuang, J., Manning, C.D., Ng, A.Y., Potts,
  C.: Recursive deep models for semantic compositionality over a sentiment
  treebank. In: Proceedings of the 2013 conference on empirical methods in
  natural language processing. pp. 1631--1642 (2013)

\bibitem{spitkovsky2011punctuation}
Spitkovsky, V.I., Alshawi, H., Jurafsky, D.: Punctuation: Making a point in
  unsupervised dependency parsing. In: Proceedings of the Fifteenth Conference
  on Computational Natural Language Learning. pp. 19--28 (2011)

\bibitem{tai2015improved}
Tai, K.S., Socher, R., Manning, C.D.: Improved semantic representations from
  tree-structured long short-term memory networks. In: Proceedings of the 53rd
  Annual Meeting of the Association for Computational Linguistics and the 7th
  International Joint Conference on Natural Language Processing (Volume 1: Long
  Papers). pp. 1556--1566 (2015)

\bibitem{vaswani2017attention}
Vaswani, A., Shazeer, N., Parmar, N., Uszkoreit, J., Jones, L., Gomez, A.N.,
  Kaiser, {\L}., Polosukhin, I.: Attention is all you need. In: Advances in
  neural information processing systems. pp. 5998--6008 (2017)

\bibitem{wang2014feature}
Wang, H., Liu, L., Song, W., Lu, J.: Feature-based sentiment analysis approach
  for product reviews. Journal of software  \textbf{9}(2),  274--279 (2014)

\bibitem{yang2016hierarchical}
Yang, Z., Yang, D., Dyer, C., He, X., Smola, A., Hovy, E.: Hierarchical
  attention networks for document classification. In: Proceedings of the 2016
  conference of the North American chapter of the association for computational
  linguistics: human language technologies. pp. 1480--1489 (2016)

\bibitem{yenicelik2020does}
Yenicelik, D., Schmidt, F., Kilcher, Y.: How does {BERT} capture semantics? a
  closer look at polysemous words. In: Proceedings of the Third Blackbox NLP
  Workshop on Analyzing and Interpreting Neural Networks for NLP. pp. 156--162
  (2020)

\bibitem{zhang2018deep}
Zhang, L., Wang, S., Liu, B.: Deep learning for sentiment analysis: A survey.
  Wiley Interdisciplinary Reviews: Data Mining and Knowledge Discovery
  \textbf{8}(4),  e1253 (2018)

\bibitem{zhang2015character}
Zhang, X., Zhao, J., LeCun, Y.: Character-level convolutional networks for text
  classification. Advances in neural information processing systems
  \textbf{28},  649--657 (2015)

\end{thebibliography}
\end{document}